# A Fast and Scalable Joint Estimator for Learning Multiple Related Sparse Gaussian Graphical Models

**Beilun Wang**
University of Virginia

**Ji Gao**
University of Virginia

**Yanjun Qi**
University of Virginia

## Abstract

Estimating multiple sparse Gaussian Graphical Models (sGGMs) jointly for many related tasks (large $K$) under a high-dimensional (large $p$) situation is an important task. Most previous studies for the joint estimation of multiple sGGMs rely on penalized log-likelihood estimators that involve expensive and difficult non-smooth optimizations. We propose a novel approach, FASJEM for fast and scalable joint structure-estimation of multiple sGGMs at a large scale. As the first study of joint sGGM using the Elementary Estimator (EE) framework, our work has three major contributions: (1) We solve FASJEM through an entry-wise manner which is parallelizable. (2) We choose a proximal algorithm to optimize FASJEM. This improves the computational efficiency from $O(Kp^3)$ to $O(Kp^2)$ and reduces the memory requirement from $O(Kp^2)$ to $O(K)$. (3) We theoretically prove that FASJEM achieves a consistent estimation with a convergence rate of $O(\log(Kp)/n_{tot})$. On several synthetic and four real-world datasets, FASJEM shows significant improvements over baselines on accuracy, computational complexity and memory costs.

## 1 Introduction

The past decade has seen a revolution in collecting large-scale heterogeneous data from many scientific fields. For instance, genomic technologies have delivered fast and accurate molecular profiling data across many cellular contexts (e.g., cell lines or stages) from national projects like ENCODE[1]. Given such data, understanding and quantifying variable graphs across multiple contexts is a fundamental analysis task. Such variable graphs can significantly simplify network-driven studies about diseases or drugs[2]. The number of contexts those applications need to consider grows extremely fast. For example, the ENCODE [1] project, being generated over ten years with contributions from bio-labs across the world, contains expression data from 147 different human cell types (i.e., the number of tasks $K = 147$) in 2016. Besides, the number of variables (denoted as $p$) is also quite large, ranging from thousands (e.g., gene) to hundreds of thousands (e.g., SNP[3]).

We formulate this data analysis problem as jointly estimating $K$ conditional dependency graphs $G^{(1)}, G^{(2)}, \ldots, G^{(K)}$ from data samples accumulated from $K$ distinct conditions. For homogeneous data samples from a given $i$-th condition, one typical approach in the literature is the sparse Gaussian Graphical Model(sGGM)[4, 5, 6]. sGGM assumes data samples are independently and identically drawn from $N_p(\mu^{(i)}, \Sigma^{(i)})$, a multivariate normal distribution with mean $\mu^{(i)}$ and covariance matrix $\Sigma^{(i)}$. The graph structure $G^{(i)}$ is encoded by the sparsity pattern of the inverse covariance matrix, also named precision matrix, $\Omega^{(i)}$. $\Omega^{(i)} := (\Sigma^{(i)})^{-1}$. In $G^{(i)}$ an edge does not connect $j$-th node and $k$-th node (i.e., conditional independent) if and only if $\Omega^{(i)}_{jk} = 0$. sGGM imposes an $\ell_1$ penalty on the parameter $\Omega^{(i)}$. For heterogeneous data samples, rather than estimating sGGM of each condition separately, a multi-task formulation that jointly estimates $K$ different but related sGGMs can lead to a better generalization[7].

Most previous studies[8, 9, 10, 11, 12, 13, 14, 15] for joint estimation of multiple sGGMs relied on optimizing $\ell_1$ regularized likelihood function plus an extra penalty function $\mathcal{R}'$. This extra regularizer $\mathcal{R}'$, which varies in different estimators, enforces similarity among multiple estimated networks. Since the penalized likelihood framework includes two regularization functions $(\ell_1 + \mathcal{R}')$, these approaches cannot avoid the steps like SVD [8] and matrix multiplication [8, 9]. Both steps need $O(Kp^3)$ time complexity for computation. Be-sides, most studies in this category require all tasks' covariance matrices to locate in the main memory[8, 9, 10] (for their optimization). Storing all elements needs $O(Kp^2)$ memory space. As a result, this category of models are difficult to scale up when the dimension $p$ or the number of tasks $K$ are large.

In this paper, we propose a novel model, namely <u>fa</u>st and <u>s</u>calable <u>j</u>oint <u>e</u>stimator for <u>m</u>ultiple sGGM (FASJEM), for estimating multiple sGGMs jointly. Briefly speaking, this paper makes the following contributions:

- **Novel approach:** FASJEM presents a new way of learning multi-task sGGMs by extending the elementary estimator [16]. (Section 3)
- **Fast optimization:** We optimize FASJEM through an entry-wise and group-entry-wise manner that can dramatically improve the time complexity to $O(Kp^2)$. (Section 3 and Section 3.3)
- **Scalable optimization:** The optimization of our estimators is scalable. We reduce the memory cost to $O(K)$ (i.e., requiring to store at most $K$ entries in the main memory). (Section 5)
- **Method variations:** We propose two variations of FASJEM: (1) FASJEM-G uses a group-2 norm to connect multiple sGGMs. (2) FASJEM-I uses a group-infinite norm to connect multiple related sGGMs. Both methods show better performance over their corresponded "Joint graphical lasso" (JGL) baselines. (Section 3 and Section 6)
- **Convergence rate:** We theoretically prove the convergence rate of FASJEM as $O(\log(Kp)/n_{tot})$. This rate shows the benefit of joint estimation, which significantly improves the convergence rate $O(\frac{\log p}{n})$ of single task sGGM (with $n$ samples). (Section 5)
- **Evaluation:** FASJEM is evaluated using several synthetic datasets and four real-world biomedical datasets. It performs better than the baselines not only on accuracy but also with respect to the time and storage requirements. (Section 6)

*Att:* Due to space limit, we have put details of certain contents (e.g., proofs) in the appendix. Notations with "S:" as prefix in the numbering mean the corresponding contents are in the appendix. For example, full proofs are in Section S:8.

*Notations:* We focus on the problem of estimating $K$ sGGMs from a $p$-dimensional aggregated dataset in the form of $K$ different data matrices. $X^{(i)}_{n_i \times p}$ describes the data matrix for the $i$-th task, which includes $n_i$ data samples described by $p$ different feature variables. The total number of data samples is $n_{tot} = \sum_{i=1}^{K} n_i$. We use notation $\Omega$ for the precision matrices and $\widehat{\Sigma}$ for the estimated covariance matrices. Given a $p$-dimensional vector $x = (x_1, x_2, \ldots, x_p)^T \in \mathbb{R}^p$, $||x||_1 = \sum_i |x_i|$ represent the $l_1$-norm of $x$. $||x||_\infty = \max_i |x_i|$ is the $l_\infty$-norm of $x$. $||x||_2 = \sqrt{\sum_i x_i^2}$, $\ell_2$-norm of $x$.

## 2 Background

**Single-task sGGM:** The classic formulation of sparse Gaussian Graphical model [6] for a single given task (or context) (single sGGM) is the "graphical lasso" estimator (GLasso) [6, 17] that solves the following penalized maximum likelihood estimation (MLE) problem:

$$\operatorname*{argmin}_{\Omega > 0} -\log\det(\Omega) + <\Omega, \Sigma> + \lambda_n ||\Omega||_1 \quad (2.1)$$

**Elementary estimator for single sGGM (EE-sGGM):** Recently the seminal study[18] generalized this formulation into a so-called M-estimator framework:

$$\operatorname*{argmin}_{\theta} \mathcal{L}(\theta) + \lambda_n \mathcal{R}(\theta) \quad (2.2)$$

where $\mathcal{R}(\cdot)$ represents a decomposable regularization function in [18] and $\mathcal{L}(\cdot)$ represents a loss function (e.g., the negative log-likelihood function $-L(\cdot)$ in sGGM). Recent studies[19, 16] propose a new category of estimators named "Elementary estimator" [1], whose solution achieves the same optimal convergence rate as Eq. (2.2) when satisfying certain conditions. These estimators have the following general formulation:

$$\operatorname*{argmin}_{\theta} \mathcal{R}(\theta)$$
$$\text{subject to:} \mathcal{R}^*(\theta - \widehat{\theta}_n) \leq \lambda_n \quad (2.3)$$

Where $\mathcal{R}^*(\cdot)$ is the dual norm of $\mathcal{R}(\cdot)$,

$$\mathcal{R}^*(v) := \sup_{u \neq 0} \frac{<u, v>}{\mathcal{R}(u)} = \sup_{\mathcal{R}(u) \leq 1} <u, v>. \quad (2.4)$$

$\widehat{\theta}_n$ represents the backward mapping of $\theta$. We provide detailed explanations of backward mapping and backward mapping for Gaussian case in the Appendix Section S:1. For sGGM, it is easy to derive $\Omega$ through the backward mapping on its covariance matrix $\Sigma$, which is $\Sigma^{-1}$. However, under the high-dimensional setting, when $p > n$, the sample covariance matrix $\widehat{\Sigma}$ is not full rank, therefore is not invertible. Thus the authors of [16] proposed a proxy backward mapping on the covariance matrix $\widehat{\Sigma}$ under high-dimensional settings as $(T_v(\widehat{\Sigma}))^{-1}$. Here $[T_v(M)]_{ij} := \rho_v(M_{ij})$ where $\rho_v(\cdot)$ is chosen to be a soft-thresholding function. [16] proves that this approximation will not change the convergence rate of sGGM estimation. Using Eq. (2.3), [16] proposed a new estimator for sGGM (the so-called elementary estimator for sGGM):

$$\operatorname*{argmin}_{\Omega} ||\Omega||_1$$
$$\text{subject to:} ||\Omega - [T_v(\widehat{\Sigma})]^{-1}||_\infty \leq \lambda_n \quad (2.5)$$

This estimator has a closed-form solution[16] and has been shown to be more practical and scalable than GLasso in large-scale settings. $v$ is a hyperparameter

---
[1] We denote this category of estimators as "elementary estimator" or "EE" in the rest of paper.

which ensures that $T_v(\widehat{\Sigma})$ is invertible.

**Multi-task sGGM (Multi-sGGM):** For joint estimation of multiple sGGMs, most studies focus on adding a second norm which enforces the group property among multiple tasks. Previous studies on the joint estimation of multiple sGGMs can be summarized using Eq. (2.6),

$$\operatorname*{argmin}_{\Omega^{(i)} \succ 0} \sum_{i=1}^{K} (-L(\Omega^{(i)})) + \lambda_n \sum_{i=1}^{K} ||\Omega^{(i)}||_1 \\ + \lambda'_n \mathcal{R}'(\Omega^{(1)}, \Omega^{(2)}, \ldots, \Omega^{(K)}) \quad (2.6)$$

where $\Omega^{(i)}$ denotes the precision matrix for $i$-th task. $\Omega^{(i)} \succ 0$ means that $\Omega^{(i)}$ needs to be a positive definite matrix. $\mathcal{R}'(\cdot)$ represents the second penalty function for multi-tasking.

**Superposition structured estimator (SS estimator):** The above Eq. (2.6) is a special case (explained in Section 3) of the following superposition structured estimators [18]:

$$\operatorname*{argmin}_{(\theta_\alpha)_{\alpha \in I}} \mathcal{L}(\sum_{\alpha \in I} \theta_\alpha) + \sum_{\alpha \in I} \lambda_\alpha \mathcal{R}_\alpha(\theta_\alpha). \quad (2.7)$$

$\{\mathcal{R}_\alpha(\cdot) | \alpha \in I\}$ are a set of regularization functions and $(\lambda_\alpha)_{\alpha \in I}$ are the regularization penalties. The target parameter is $\theta = \sum_{\alpha \in I} \theta_\alpha$, a superposition of $\theta_\alpha$.

**Elementary superposition-structured moment estimator (ESS moment estimator):** Similar to Eq. (2.3), a recent study[20] extends the elementary estimator for sparse covariance matrices to the case of superposition-structured moments and named this extension as "Elem-Super-Moment" (ESM) estimator.[2]

$$\operatorname*{argmin}_{\theta_1, \theta_2, \ldots, \theta_{|I|}} \sum_{\alpha \in I} \lambda_\alpha \mathcal{R}_\alpha(\theta_\alpha) \\ \text{Subject to: } \mathcal{R}_\alpha^*(\widehat{\theta} - \sum_{\alpha \in I} \theta_\alpha) \leq \lambda_\alpha \quad \forall \alpha \in I. \quad (2.8)$$

## 3 Method: A fast and scalable joint estimator for multi-sGGM

The penalized likelihood framework for multi-task sGGMs in Eq. (2.6) involves a hybrid of two regularization functions ($\ell_1 + \mathcal{R}'$). Studies in this direction cannot avoid the expensive steps like SVD and matrix multiplication and also require to store $K$ covariance matrices in the main memory. Since this paper aims to design a scalable joint estimator for multi-sGGM under large-scale settings, extending the elementary estimator of single-task sGGM [16] to multi-task formulation becomes a natural choice.

For multi-task sGGMs, we can denote that $\Omega_{tot} = (\Omega^{(1)}, \Omega^{(2)}, \ldots, \Omega^{(K)})$ and $\Sigma_{tot} = (\Sigma^{(1)}, \Sigma^{(2)}, \ldots, \Sigma^{(K)})$. $\Omega_{tot}$ and $\Sigma_{tot}$ are both $p \times Kp$ matrices (i.e., $Kp^2$ parameters to estimate). Now define an inverse function as $\operatorname{inv}(A_{tot}) := (A^{(1)^{-1}}, A^{(2)^{-1}}, \ldots, A^{(K)^{-1}})$, where $A_{tot}$ is a given $p \times Kp$ matrix with the same structure as $\Sigma_{tot}$. Furthermore, we add a new hyperparameter variable $\epsilon = \frac{\lambda'_n}{\lambda_n}$.

Let $I = \{1, 2\}$ and $\theta_1 = \theta_2 = \frac{1}{2} \Omega_{tot}$. We can clearly tell that Eq. (2.6) is a special case of the superposition structured estimation in Eq. (2.7). The ESS (elementary superposition structured) moment estimator (Eq. (2.8)) extends the elementary estimator of structured covariance matrix to elementary superposition-structured estimator for estimating covariance matrices with a hybrid structure (e.g., sparse + low rank). This motivates us to propose the following elementary superposition estimator for learning multi-task sGGM:

$$\operatorname*{argmin}_{\Omega_{tot}} ||\Omega_{tot}||_1 + \epsilon \mathcal{R}'(\Omega_{tot}) \\ s.t. ||\Omega_{tot} - \operatorname{inv}(T_v(\widehat{\Sigma}_{tot}))||_\infty \leq \lambda_n \quad (3.1) \\ \mathcal{R}'^*(\Omega_{tot} - \operatorname{inv}(T_v(\widehat{\Sigma}_{tot}))) \leq \epsilon \lambda_n$$

Here $|| \cdot ||_1^* = || \cdot ||_\infty$ (the dual norm of $l_1$-norm is $l_\infty$-norm). $\mathcal{R}'(\cdot)$ represents a regularizer on $\Omega_{tot}$ to enforce that $\{\Omega^{(i)}\}$ share certain similarity. $\mathcal{R}'^*(\cdot)$ is the dual norm of $\mathcal{R}'(\cdot)$. We name this novel formulation as FASJEM. By varying $\mathcal{R}'(\cdot)$, we can get a variety of FASJEM estimators.

Section 5 theoretically proves the convergence rate of FASJEM as $O(\log(Kp)/n_{tot})$. Our theory proof is inspired by the ESS moment estimator [20], the SS estimator [21] and the EE-sGGM [16].

### 3.1 Method I: FASJEM-G

For multi-task regularization, the first $\mathcal{R}'(\cdot)$ we try is the $\mathcal{G}, 2$-norm (i.e., $\mathcal{R}'(\cdot) = || \cdot ||_{\mathcal{G},2}$). This norm is inspired by JGL-group lasso[8]. $\mathcal{G}, 2$-norm constrains the parameters in the same group to have the same level of sparsity. In multi-task sGGMs, group set $\mathcal{G} := \{g_{j,k}\}$, where $g_{j,k} = \{\Omega_{j,k}^{(i)} | i = 1, \ldots, K\}$. Suppose $g$ is an arbitrary group in group set $\mathcal{G}$ and totally we have $p^2$ groups. $||\Omega_{tot}||_{\mathcal{G},2} = \sum_{j=1}^{p} \sum_{k=1}^{p} ||(\Omega_{j,k}^{(1)}, \Omega_{j,k}^{(2)}, \ldots, \Omega_{j,k}^{(i)}, \ldots, \Omega_{j,k}^{(K)})||_2$. When $\mathcal{R}'(\cdot) = || \cdot ||_{\mathcal{G},2}$, we name Eq. (3.1) as FASJEM-G (short form of FASJEM-Group2). We solve FASJEM-G using a parallel proximal based optimization formulation from[22]. Algorithm 1 summarizes the detailed optimization steps and the four proximity operators implemented on GPU are listed in Table 1[3]. The optimization sequence of Algorithm 1 converges Q-linearly (See Eq. (S:2–10)).

---

[2][20] has proved that this class of ESM estimators achieves the same convergence rate as the corresponding estimators (with the same superposition of structures) using the penalized MLE formulation under certain conditions.

[3]The non-GPU version of the four proximity operators are in Eq. (S:2–2) to Eq. (S:2–5).

## 3.2 Method II: FASJEM-I

As shown in Section 4, most previous models for multi-task sGGMs varied the second norm $R'$ to obtain different models. Similarly we can easily change $R'(\cdot)$ in Eq. 3.1 into any other desired norm to extend our FASJEM. For instance, we can change $R'(\cdot)$ to group-infinity norm $||\cdot||_{\mathcal{G},\infty}$.

$$||\Omega_{tot}||_{\mathcal{G},\infty} = \sum_{j=1}^{p}\sum_{k=1}^{p}||(\Omega_{j,k}^{(1)},\Omega_{j,k}^{(2)},\ldots,\Omega_{j,k}^{(i)},\ldots,\Omega_{j,k}^{(K)})||_{\infty}.$$

This norm is inspired by a multi-task sGGM proposed by [11]. When using group-infinity norm, we get FASJEM-I(short for FASJEM-Groupinf). We can derive the optimization for FASJAM-I by changing two proximities in Algorithm 1. Considering that the original formulation in [11] is similar with JGL[8], in the rest of this paper, we call the model from [11] as JGL-groupinf or JGL-I (the corresponded baseline for FASJEM-I).

## 3.3 Proximal Algorithm for Optimization

Eq. (3.1) includes a convex programming task since the norms we choose are convex. By simplifying notations and adding another parameter, we reformulate it to:

$$\underset{\theta_1,\theta_2}{\operatorname{argmin}} f_1(\theta_1) + f_2(\theta_2)$$

$$\text{subject to :}||\theta_1 - \operatorname{inv}(T_v(\widehat{\Sigma}_{tot}))||_{\infty} \leq \lambda_n \quad (3.2)$$

$$\mathcal{R}'^*(\theta_2 - \operatorname{inv}(T_v(\widehat{\Sigma}_{tot}))) \leq \epsilon\lambda_n$$

$$\theta_1 = \theta_2$$

Where $f_1(\cdot) = ||\cdot||_1$ and $f_2(\cdot) = \epsilon||\cdot||_{\mathcal{G},2}$. Then we convert Eq. (3.2) to the following equivalent and distributed formulation:

$$\underset{\theta_1,\theta_2,\theta_3,\theta_4}{\operatorname{argmin}} f_1(\theta_1) + f_2(\theta_2) + f_3(\theta_3) + f_4(\theta_4) \quad (3.3)$$

$$\text{subject to: } \theta_1 = \theta_2 = \theta_3 = \theta_4$$

Here $f_3(\theta) = \mathcal{I}_{\{||\theta-inv(T_v(\Sigma_{tot}))||_{\infty}\leq\lambda_n\}}(\theta)$ and $f_4(\theta) = \mathcal{I}_{\{||\theta-inv(T_v(\Sigma_{tot}))||_{\mathcal{G},2}^*\leq\epsilon\lambda_n\}}(\theta)$. $\mathcal{I}_C(\cdot)$ represents the indicator function of a convex set $C$ as $\mathcal{I}_C(x) = 0$ when $x \in C$. Otherwise $\mathcal{I}_C(x) = \infty$. To solve Eq. (3.3), we choose a parallel proximal based algorithm[22] summarized in Algorithm 1. Besides the distributed nature, the proximal algorithm also bring in the benefit that many proximity operators are entry-wise operators for the targeted parameters. The four proximal operators for four functions $\{f_1, f_2, f_3, f_4\}$ (for CPU platform implementation) are included in the Equations Eq. (S:2–2) to Eq. (S:2–5) in Section S:2. With the benefits as proximal operators, Eq. (S:2–2) and Eq. (S:2–4) are entry-wise and Eq. (S:2–3) and Eq. (S:2–5) are group entry-wise.

## 3.4 GPU Implementation of FASJEM-G

We further revise Algorithm 1 to take advantage of the advanced computational architecture–GPU. This algorithm cannot be directly parallelized on GPU, because GPU is slow for multiple branches based predictions.

Table 1: Four proximity operators implemented on GPU platform.

| | |
|---|---|
| $[\operatorname{prox}_{\gamma f_1}(x)]_{j,k}^{(i)}$ | $\max((x_{j,k}^{(i)}-\gamma),0)+\min(0,(x_{j,k}^{(i)}+\gamma))$ |
| $\operatorname{prox}_{\gamma f_2}(x_g)$ | $x_g\max((1-\frac{\gamma}{||x_g||_2}),0)$ |
| $[\operatorname{prox}_{\gamma f_3}(x)]_{j,k}^{(i)}$ | $\min(\max(x_{j,k}^{(i)}-a_{j,k}^{(i)},-\lambda_n),\lambda_n)+a_{j,k}^{(i)}$ |
| $\operatorname{prox}_{\gamma f_4}(x_g)$ | $\max(\frac{\lambda_n}{||x_g-a_g||_2},1)(x_g-a_g)+a_g$ |

Therefore, we convert those four operators prox(·) in Algorithm 1 into single soft-threshold based operators which only include simple algorithmic operations like + or max. These operators can be easily parallelized on GPU[23]. The four proximity operators we use to implement FASJEM-G on GPU are summarized in Table 1. More details are included in Section S:2.

**Algorithm 1** Parallel proximal algorithm[4]
---
**input** $K$ given data blocks $X^{(1)}, X^{(2)}, \ldots, X^{(K)}$. Hyperparameter:$\alpha, \epsilon, v, \lambda_n$ and $\gamma$. Learning rate: $0 < \rho < 2$. Max iteration number $iter$.
**output** $\Omega_{tot}$
1: Compute $\Sigma_{tot}$ from $X^{(1)}, X^{(2)}, \ldots, X^{(K)}$
2: Initialize $\theta^0 = \operatorname{inv}(T_v(\Sigma_{tot})), \theta_j^0 = \operatorname{inv}(T_v(\Sigma_{tot}))$ for $j \in \{1,2,3,4\}$ and $a = \operatorname{inv}(T_v(\Sigma_{tot}))$.
3: **for** $i = 0$ **to** $iter$ **do**
4:   $p_1^i = \operatorname{prox}_{4\gamma f_1}\theta_1^i$
5:   $p_2^i = \operatorname{prox}_{4\gamma f_2}\theta_2^i$
6:   $p_3^i = \operatorname{prox}_{4\gamma f_3}\theta_3^i$
7:   $p_4^i = \operatorname{prox}_{4\gamma f_4}\theta_4^i$
8:   $p^i = \frac{1}{4}(\sum_{j=1}^{4}\theta_j^i)$
9:   **for** $j = 1,2,3,4$ **do**
10:     $\theta_j^{i+1} = \theta_j^i + \rho(2p^i - \theta^i - p_j^i)$
11:   **end for**
12:   $\theta^{i+1} = \theta^i + \rho(p^i - \theta^i)$
13: **end for**
14: $\Omega_{tot} = \theta^{iter}$
**output** $\Omega_{tot}$

## 4 Connecting to Relevant Studies

**Estimators for Single task sGGM:** Sparse GGM is a highly active topic in the recent literature. Roughly speaking, there exist three major categories of estimators for sGGM. A key class of estimators is based on the regularized maximum likelihood optimization. The popular estimator "graphical lasso"(GLasso) considers maximizing a $\ell_1$ penalized normal likelihood [6, 17, 24, 25]. As the second type, CLIME estimator[26] learns sGGM by solving a constrained $\ell_1$ optimization. The CLIME formulation can be converted into multiple subproblems of linear programming and has shown more favorable theoretical properties than GLasso. The linear programming while convex, is computationally expensive for large-scale tasks. Recently as a third group of stud-

---
[4]Four proximity operators used on GPU are defined in Table 1. Hyperparameters are explained in Section 6. Here $j,k = 1,\ldots,p, i = 1,\ldots,K$ and $g \in \mathcal{G}$.

Beilun Wang, Ji Gao, Yanjun Qiies, soft-thresholding based elementary estimators[16] have been introduced for inferring undirected sparse Graphical models. This paper mostly follows the third category. Besides, there exists quite a number of recent studies trying to scale up single sGGM to a large scale. For example, the BigQUIC algorithm [27] proposes an asymptoticly quadratic optimization to estimate sGGM. The elementary estimator [16] of sGGM is an advanced version of BigQUIC.

**Multi-sGGM: Previous Likelihood based Estimators.** Most previous methods to jointly estimate multiple sGGMs can be formulated using the penalized MLE Eq. (2.6), including for instance, (1) Joint graphical lasso (JGL-group uses $\mathcal{G}, 2$-norm in Section 3.1 [8]), (2) Node-perturbed JGL [9], (3) Simone [10], and (4) multi-task sGGM proposed by [11]. These methods differ with respect to the second regularization function $\mathcal{R}'(\cdot)$ they used to enforce their assumption of similarity among tasks.

**Optimization and Computational Comparison:** We use JGL-group and the model proposed by [11] (we name it as JGL-groupInf) as baselines in our experiments. As we mentioned in Section 1, the bottleneck of optimizing multi-sGGM in JGL-group is the step of SVD that needs $O(Kp^3)$ time complexity and requires storing $K$ covariance matrix ($O(Kp^2)$ memory cost). Differently, JGL-Groupinf chose a coordinate descent method and proved that their optimization is equivalent to $p$ sequences of quadratic subproblems, each of which costs $O(K^3p^3)$ computation. Therefore the total computational complexity of JGL-Groupinf is $O(K^3p^4)$. Besides, this coordinate descent method needs to store all $K$ covariance matrices in the main memory ($O(Kp^2)$ memory cost). Table 2 compares our model with two baselines in terms of time and space cost. Solving our model relies totally on entry-wise and group-entry-wise procedures. Its time complexity is $O(Kp^2)$. This is much faster than the baselines, especially in high-dimensional settings (Table 2). [5] Moreover, in our optimization, learning the parameters for each group $\{\Omega_{j,k}^{(i)}|i=1,\ldots,K\}$ does not rely on other groups. This means we only need to store $K$ entries of the same group in the memory for computing Eq. (S:2–4) and Eq. (S:2–5). The space complexity $O(K)$ is much smaller than previous methods' $O(Kp^2)$ requirement. [6] The comparisons are in Table 2.

Table 2: Comparison to Previous multi-sGGM methods

| References | Computational Complexity | Memory Cost |
|---|---|---|
| JGL-Group [8] | $O(Kp^3)$ | $O(Kp^2)$ |
| JGL-GroupInf [11] | $O(K^3p^4)$ | $O(Kp^2)$ |
| FASJEM Models | $O(Kp^2)$ (if paralleling completely, $O(K)$ ) | $O(K)$ |

**Previous Studies using Elementary based Estimators:** Most previous studies of multi-sGGMs follow the penalized MLE framework. Few works of Multi-task sGGM follow the CLIME formulation, since it is not easy to transfer two regularizers into the CLIME formulation (summarized in Table S:1). Based on the authors' knowledge, no previous multi-sGGM studies have followed the elementary estimators(EE) formulation. As a simple soft-thresholding based estimator, elementary estimators (EE) have been used for other tasks as well. Table S:1 summarizes three different types of previous tasks for which EE can be applied: high-dimensional regression, single sGGM and multi-sGGM. For comparison, we show how these tasks have been solved through the penalized likelihood framework in the second column and use the the third column to show studies following the CLIME formulation.

**Convergence Rate Analysis:** Although previous joint sGGMs work well on datasets whose $K$ and $p$ are relatively small, two important questions remain unanswered: (1) what's the statistical convergence rate of these joint estimators? and (2) what's the benefits of joint learning? The convergence rate of estimating single-task sGGM has been well investigated[6, 17, 24, 25]. These studies proved that the estimator of single-task sGGM holds a consistent convergence rate $O(\sqrt{\frac{\log p}{n}})$ if given $n$ data samples. However, none of the previous joint-sGGM studies provide such theoretical analysis. Experimental evaluations in previous joint-sGGM papers have shown better performance of running joint estimators over running single-task sGGM estimators on each dataset separately. However, it hasn't been proven that theoretically this joint estimation is better. We successfully answer these two remaining questions in Section 5.

## 5 Theoretical Analysis

In this section, we prove that our estimator can be optimized asynchronously in a group entry-wise manner. We also provide the proof of the theoretical error bounds of FASJEM.

---

[5] Note that the discussion of time complexity is for each iteration in optimization. We show the Q-linear convergence for all first-order multi-task sGGM estimators in Eq. (S:2–10). Since the baselines and our methods all use first-order optimization, we assume the number of iterations is the same among all methods.

[6] We have provided a GPU implementation of FASJEM in Section 3.3. Although SVD or matrix inversion can also be speed up by GPU parallelization, these method cannot avoid the $O(Kp^2)$ memory cost, which is a huge bottleneck for large-scale problems. In Section 5 we prove that our estimator is completely group entry-wise and asynchronously optimizable, this makes FASJEM only require $O(K)$ memory storage.

# A Fast and Scalable Joint Estimator for Learning Multiple Related Sparse Gaussian Graphical Models

## 5.1 Group entry-wise and parallelizing optimizable

**Theorem 5.1.** *(FASJEM is Group entry-wise optimizable) Suppose we use FASJEM to infer multiple inverse of covariance matrices summarized as $\widehat{\Omega}_{tot}$. $\{\widehat{\Omega}^{(i)}_{j,k}|i=1,\ldots,K\}$ describes a group of $K$ entries at $(j,k)$ position. Varying $j \in \{1,2,\ldots,p\}$ and $k \in \{1,2,\ldots,p\}$, we have totally $p \times p$ groups. If these groups are independently estimated by FASJEM, then we have,*

$$\bigcup_{j=1}^{p} \bigcup_{k=1}^{p} \{\widehat{\Omega}^{(i)}_{j,k}|i=1,\ldots,K\} = \widehat{\Omega}_{tot}. \quad (5.1)$$

*Proof.* Eq. (S:2–6) and Eq. (S:2–8) are soft-thresholding based operators on each entry. Eq. (S:2–7) and Eq. (S:2–9) are soft-thresholding operators on each group of entries. □

**Corollary 5.2.** *We can decompose FASJEM into $p \times p$ subproblems that are independent from each other, and solve each subproblem at a time. Therefore our estimator only requires $O(K)$ memory storage for computation.*

This corollary proves the claims we showed in section 4. Through Theorem (5.1), it is important to notice that the optimization on multiple groups of entries can be totally **parallelized**.

## 5.2 Theoretical error bounds

In this subsection, we first provide the error bounds for elementary super-position estimator (ESS estimator) under $I = \{1, 2\}$. We then use this general bound to prove the error bound for FASJEM-G. All the proofs are included in Section S:8. We also include the error bounds for elementary estimator (EE) in Section S:7.

**Extension to ESS:** For the multiple-task case, we need to consider two or more regularization functions. For instance, in FASJEM-G we assume the sparsity of parameter and the group sparsity among tasks. Since we only consider the models with two regularization function, we consider the error bounds of the following elementary super-position estimator formulation in the rest of the section.

$$\operatorname*{argmin}_{\theta_1, \theta_2} \lambda_1 \mathcal{R}_1(\theta_1) + \lambda_2 \mathcal{R}_2(\theta_2) \quad (5.2)$$

subject to: $\mathcal{R}^*_i(\widehat{\theta}_n - (\theta_1 + \theta_2)) \leq \lambda_i$, $i = 1, 2$

This equation restricts the number of penalty functions to 2. Similar to the single-task error bounds (in Section S:7), we naturally extend condition **(C2)** to the following condition:

**(C3)** $\operatorname{proj}_{\mathcal{M}^\perp_i}(\theta^*_i) = 0$, $i = 1, 2$.

We borrow the following condition from [21], which is a structural incoherence condition ensuring that the non-interference of different structures.

**(C4)** Let $\Phi := \max\{2 + \frac{3\lambda_1 \Psi_1(\bar{\mathcal{M}}_1)}{\lambda_2 \Psi_2(\bar{\mathcal{M}}_2)}, 2 + \frac{3\lambda_2 \Psi_2(\bar{\mathcal{M}}_2)}{\lambda_1 \Psi_1(\bar{\mathcal{M}}_1)}\}$.
$\max\{\sigma_{\max}(\mathcal{P}_{\bar{\mathcal{M}}_1}\mathcal{P}_{\bar{\mathcal{M}}_2}),$
$\sigma_{\max}(\mathcal{P}_{\bar{\mathcal{M}}_1}\mathcal{P}_{\bar{\mathcal{M}}^\perp_2})\sigma_{\max}(\mathcal{P}_{\bar{\mathcal{M}}^\perp_1}\mathcal{P}_{\bar{\mathcal{M}}^\perp_2})\} \leq \frac{1}{16\Phi^2}$

where $\mathcal{P}_{\bar{\mathcal{M}}}$ is the matrix corresponding to the projection operator for the subspace $\bar{\mathcal{M}}$. The definition of $\Psi(\cdot)$ are included in Definition (S:7.1).

With these two conditions, we have the following theorem:

**Theorem 5.3.** *Suppose that the true parameter $\theta^*$ satisfies the conditions **(C3)(C4)** and $\lambda_i \geq \mathcal{R}^*_i(\widehat{\theta} - \theta^*)$, then the optimal point $\widehat{\theta}$ of Eq. (5.2) has the following error bounds:*

$$\mathcal{R}^*_i(\widehat{\theta} - \theta^*) \leq 2\lambda_i, \ i = 1, 2 \quad (5.3)$$

$$\mathcal{R}_i(\widehat{\theta} - \theta^*) \leq \frac{32}{\lambda_i}(\max_i \lambda_i \Psi(\bar{\mathcal{M}}_i))^2, \ i = 1, 2 \quad (5.4)$$

$$||\widehat{\theta} - \theta^*||_F \leq 8 \max_i \lambda_i \Psi(\bar{\mathcal{M}}_i) \quad (5.5)$$

Notice that for FASJEM-G model, $\mathcal{R}_1 = ||\cdot||_1$ and $\mathcal{R}_2 = ||\cdot||_{\mathcal{G},2}$. Based on the results in[18], $\Psi(\mathcal{M}_1) = \sqrt{s}$ and $\Psi(\mathcal{M}_2) = \sqrt{s_{\mathcal{G}}}$, where $s$ is the number of nonzero entries in $\Omega_{tot}$ and $s_g$ is the number of groups in which there exists at least one nonzero entry. Clearly $s > s_g$. Also in practice, to utilize group information, we have to choose hyperparameter $\lambda_n > \lambda'_n$ ($\lambda_1 > \lambda_2$ in Eq. (5.5)). Therefore by Theorem (5.3), we have the following theorem,

**Theorem 5.4.** *Suppose that $\mathcal{R}_1 = ||\cdot||_1$ and $\mathcal{R}_2 = ||\cdot||_{\mathcal{G},2}$ and the true parameter $\Omega^*_{tot}$ satisfies the conditions **(C3)(C4)** and $\lambda_i \geq \mathcal{R}^*_i(\widehat{\Omega}_{tot} - \Omega^*_{tot})$, then the optimal point $\widehat{\Omega}_{tot}$ of Eq. (3.1) has the following error bounds:*
$||\widehat{\Omega}_{tot} - \Omega^*_{tot}||_F \leq 8\sqrt{s}\lambda_n$.

We then derive a corollary of Theorem (5.4) for FASJEM-G. A prerequisite is to show that $inv(T_v(\widehat{\Sigma}_{tot}))$ is well-defined. The following conditions define a broad class of sGGM that satisfy the requirement. Similar results are also introduced by[16].

**Conditions for elementary estimator of sGGM:**
**C-MinInf$\Sigma$** The true parameter $\Omega^*_{tot}$ of Eq. (5.2) has bounded induced operator norm, i.e., $|||\Omega^{(i)*}|||_\infty := \sup_{w \neq 0 \in \mathbb{R}^p} \frac{||\Sigma^{(i)*}w||_\infty}{|w|_\infty} \leq \kappa_1 \forall i$.

**C-Sparse$\Sigma$** The true multiple covariance matrices $\Sigma^*_{tot} := inv(\Omega^*_{tot})$ are "approximately sparse" along the lines [28] : for some positive constant $D$, $\Sigma^{(i)*}_{j,j} \leq D$ for all diagonal entries. Moreover, for some $0 \leq q < 1$ and $c_0(p)$, $\max_i \sum_{j=1}^{p} |\Sigma^{(i)*}_{j,k}|^q \leq c_0(p) \forall i$. If $q = 0$, then this condition reduce to $\Sigma^*$ being sparse. We additionally require $\inf_{w \neq 0 \in \mathbb{R}^p} \frac{|\Omega^{(i)*}w|_\infty}{|w|_\infty} \geq \kappa_2$.

**Error bounds of FASJEM-group:** In FASJEM, $\theta_1^* = \theta_2^* = \frac{1}{2}\theta^*$. $\theta_i$ is the parameter w.r.t a subspace pair$(\mathcal{M}_i, \bar{\mathcal{M}}_i^\perp)$, where $i = 1, 2$.

Here $\mathcal{R}_1 = ||\cdot||_1$ and $\mathcal{R}_2 = ||\cdot||_{\mathcal{G},2}$. We assume the true parameter $\theta^*$ satisfies **C-MinInf**$\Sigma$ and **C-Sparse**$\Sigma$ conditions. Using the above theorems, we have the following corollary:

**Corollary 5.5.** *If we choose hyperparameters $\lambda_n' < \lambda_n$. Let $v := a\sqrt{\frac{\log p'}{n_{tot}}}$ for $p' = \max(Kp, n_{tot})$. Then for $\lambda_n := \frac{4\kappa_1 a}{\kappa_2}\sqrt{\frac{\log p'}{n_{tot}}}$ and $n_{tot} > c \log p'$, with a probability of at least $1 - 2C_1 \exp(-C_2 Kp \log(Kp))$, the estimated optimal solution $\widehat{\Omega}_{tot}$ has the following error bound:*
$||\widehat{\Omega}_{tot} - \Omega_{tot}^*||_F \leq 32\frac{4\kappa_1 a}{\kappa_2}\sqrt{\frac{s \log p'}{n_{tot}}}\}$
*where $a$, $c$, $\kappa_1$ and $\kappa_2$ are constants.*

The convergence rate of single-task sGGM is $O(\log p/n_i)$. In high-dimensional setting, $p' = Kp$ since $Kp > n_{tot}$. Assuming $n_i = \frac{n_{tot}}{K}$, the convergence rate of single sGGM is $O(K \log p/n_{tot})$. Clearly, since $K \log p > \log(Kp)$, the convergence rate of FASJEM is better than single-task sGGM.

## 6 Experiment

Multiple simulated datasets and four real-world biomedical datasets are used to evaluate FASJEM.

### 6.1 Experimental Settings

**Baseline:** We compare (1)FASJEM-G versus JGL-group [8]; (2)FASJEM-I versus JGL-groupinf [11]. This is because the specific FASJEM estimator and its baseline share the same second-penalty function. [7] Three evaluation metrics are used for such comparisons.

- **Precision:** We use the edge-level false positive rate (FPR) and true positive rate (TPR) to measure the predicted graphs versus true graph. Repeating the process 10 times, we obtain average metrics for each method we tests. Here, FPR = $\frac{FP}{FP + TN}$ and TPR = $\frac{TP}{TP + FN}$. TP (true positive) and TN (true negative) mean the number of true nonzero entries and the number of true zero entries estimated by the predicted precision matrices. The FPR vs. TPR curve shows multi-point performance of a method over a range of the tuning parameter. The bigger the area under a FPR-TPR curve, the better a method has achieved overall.
- **Speed:** The time (log(second)) between the whole program's start and end indicates the speed of a certain method under a specific configuration of hyperparameters. To be fair, we set up two types of comparisons. The first one fixes the number of tasks ($K$) but varies the dimension ($p$). This shows the performance of each method under a high-dimensional setting. The other type fixes the dimension ($p$) but varies the number of tasks ($K$). This measures the performance of each method when having a large number of tasks.
- **Memory:** For each method, we vary the number of tasks ($K$) and the dimension ($p$) until a specific method terminates due to the "out of memory" error. This measures the memory capacity of the corresponding method.

**Our implementation:** We implement FASJEM on two different architectures: (1)CPU only and (2)GPU [8]. Similar to the JGL-group from [8], we implement the CPU version FASJEM-G and FASJEM-I with R. We choose torch7 [29] (LUA based) to program FASJEM on GPU machine. [9]

**Selection of hyper-parameters:** In this experiment, we need to choose the value of three hyper-parameters. The first one $v$ is unique for elementary-estimator based sGGM models. The second $\lambda_n$(in some models also noted as $\lambda_1$) is the main hyper-parameter we need to tune. The third $\epsilon$ equals to $\frac{\lambda_n'}{\lambda_n}$ (The notation $\lambda_2$ is normally used in related works instead of $\lambda_n'$).

- $v$: We pre-choose $v$ in the set $\{0.001i | i = 1, 2, \ldots, 1000\}$ to guarantee $T_v(\Sigma_{tot})$ is invertible.
- $\lambda_n$[10]: Recent research studies from [18] and [16] conclude that the regularization parameter $\lambda_{n_i}$ of a single task with $n_i$ samples should be chosen with $\lambda_{n_i} \propto \sqrt{\frac{\log p}{n_i}}$. Combining this result and our convergence rate analysis in Section 5, we choose $\lambda_n = \alpha\sqrt{\frac{\log Kp}{n_{tot}}}$ where $\alpha$ is a hyper-parameter. The hyperparameter $\gamma$ in Algorithm 1 equals to $\lambda_n$.
- $\epsilon$: We select the best $\epsilon$ from the set $\{0.1i | i = 1, 2, \ldots, 10\}$ using cross-validation.

### 6.2 Experiments on simulated datasets

Using the following "Random Graph Model"(RGM), we first generate a set of synthetic multivariate Gaussian datasets, each of which includes samples of $K$ tasks described by $p$ variables. From [25], this "Random Graph Model" assumes $\Omega^{(i)} = B^{(i)} + \delta^{(i)}I$, where each off-diagonal entry in $B^{(i)}$ is generated independently, equals to 0.5 with probability $0.05i$ and, equals to 0 with probability $1 - 0.05i$. $\delta^{(i)}$ is selected large enough to guarantee the positive definiteness of precision matrix.

For each case of $p$, we use this model to generate $K$ random sparse graphs. For each graph (task), $n = p/2$

---

[7]Since single-sGGM EE has a closed-form solution (i.e., no iterative steps are needed in optimization), we do not include it as baseline.

[8]Information of Experiment Machines: The machine that we use for experiments includes Intel(R) Core(TM) i7-3770 CPU @ 3.40GHz with a 8GB memory. The GPU that we use for experiments is Nvidia Tesla K40c with 2880 cores and 12GB memory.

[9]Though the ideal memory requirement of FASJEM is only $O(K)$, IO costs should also be taken into account in real implementations. As being proved, $\Omega_{tot}$ is group-entry-wise optimizable. The parameter groups are independently estimated in the parallelized style. When implementing FASJEM in a single machine (our experimental setting), we prefer to choose smaller $m$ to make full use of the main memory, where $m$ is the number of parameter groups which are estimated at the same time.

[10]$\lambda_n = 0.1$ used for time and memory experiments

A Fast and Scalable Joint Estimator for Learning Multiple Related Sparse Gaussian Graphical Models

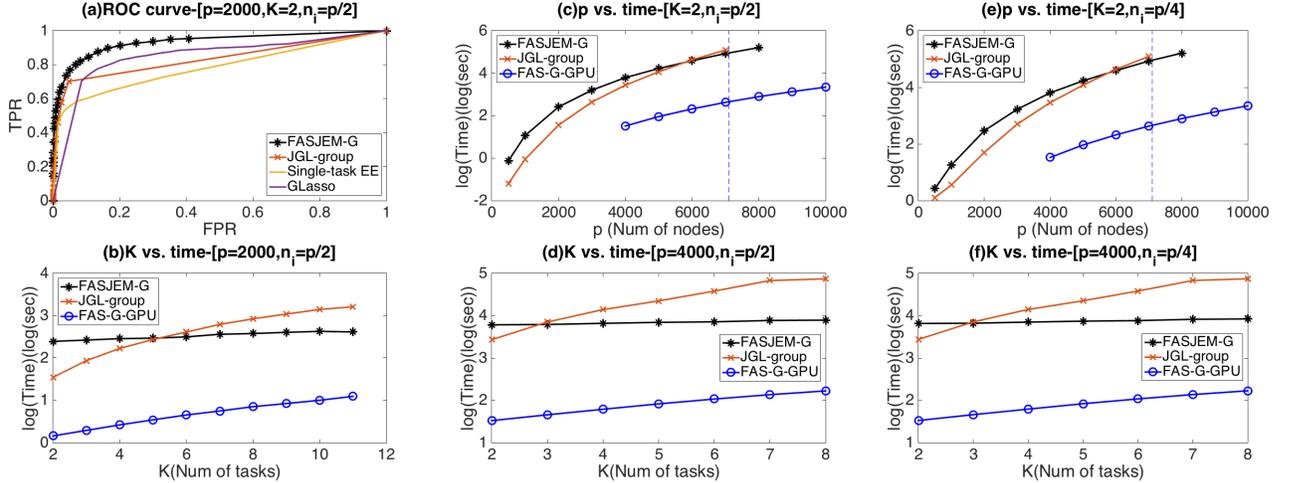

Figure 1: FASJEM-G versus JGL-group with respect to accuracy, speed and memory capacity. **(a):** FPR-TPR curves of two methods and two single-sGGM baselines on the simulated dataset using Random Graph Model when $p = 2000$ and $K = 2$. (AUC number–FASJEM-G:0.9332, JGL-group:0.5803, EE for sGGM:0.7852, GLasso:0.8504) **(c) and (e):** Time versus $p$(the number of variables) curves from FASJEM-G, JGL-group and FASJEM-G's GPU implementation. (c) uses $n_i = p/2$ and (e)$n_i = p/4$. **(b), (d) and (f):** the time versus $K$(the number of tasks) curves for two methods plus FASJEM-G-GPU. (b) uses $p = 2000$ and $n_i = p/2$, (d) uses $p = 4000$ and $n_i = p/2$ and (f) uses $p = 4000$ and $n_i = p/4$.

data samples are generated randomly by following $N(0, (\Omega^{(i)})^{-1})$. For each $(K, p)$ parameter setting we test in the experiment, we use this RGM process to generate 10 different datasets (with different random seeds). Then we apply our methods and baseline methods on these datasets to obtain estimated sGGM networks. All results or curves we show in the rest of this section are average scores/curves over 10 trials for each case of parameter configuration.

Figure 1(a) and S:2(a) present FPR vs. TPR curves of two proposed methods: FASJEM-G and FASJEM-I versus their corresponding baselines: JGL-group and JGL-groupinf, on the simulated datasets. We choose $p = 2000$ and $K = 2$. FPR-TPR curve plots are obtained by varying its tuning parameter $\lambda_n$ over a range of $\{0.05 \times \sqrt{\frac{\log Kp}{n_{tot}}} \times i | i \in \{1, 2, 3, \ldots, 30\}\}$ and interpolating the obtained performance points (We pre-choose $v$ and $\epsilon$ and the methods are introduced in Section 6.1). The two subfigures of "ROC curve" clearly show that FASJEM-G and FASJEM-I obtain better under-plot areas than corresponding JGL-group and JGL-groupinf.

Then in Figure 1(c)(e) and S:2(c)(e) we show the curves of Time vs. Dimension $p$ comparing FASJEM-G and FASJEM-I versus their baselines. Sub-figure 1(c) and S:2(c) choose $n_i = p/2$. Sub-Figures 1(e) and S:2(e) use $n_i = p/4$. The CPU curves are obtained by varying $p$ in the set of $\{1000i | i = 0.5, 1, 2, 3, \ldots, 8\}$. GPU curves are obtained by varying $p$ in the set of $\{1000i | i = 4, 5, 6, \ldots, 10\}$. The subfigure (c) "$p$ versus time-$[K = 2, n_i = p/2]$" and subfigure (e) "$p$ versus time-$[K = 2, n_i = p/4]$" in Figure 1 show that though JGL-group obtains a slightly better performance than our method under lower-dimension cases, when reaching high dimensional stages, FASJEM-G performs similarly and trains much faster than the baseline method. Figure S:2(c) and Figure S:2(e) provide similar conclusions for FASJEM-I vs JGL-groupinf. In addition, the baselines cannot handle $p \geq 8000$ because these approaches require too much memory. Clearly our proposed FASJEM methods can still perform reasonable well for the large-scale cases. This shows that our methods makes better usage of memory. Moreover, both FASJEM-G-GPU and FASJEM-I-GPU implementations spend only $\frac{1}{10}$ of train time against its CPU implementations. This proves that GPU parallelization can speed up FASJEM significantly.

Figure 1(b)(d)(f) and S:2(b)(d)(f) show the curves about "Time vs. Number of tasks-$K$" comparing our methods FASJEM-G and FASJEM-I versus two baseline methods JGL-group and JGL-groupinf respectively. These sub-figures use the varying $K$ as the x-axis over a range of $\{2, 3, \ldots, 8\}$. Sub-figures (b) use $p = 2000$, $n_i = p/2$, sub-figures (d) use $p = 4000$, $n_i = p/2$ and sub-figures (f) choose $p = 4000$, $n_i = p/4$. These figures show that the JGL-group and JGL-groupinf obtain a slightly better speed than two FASJEM, under small $K$ cases. For larger $K$, our methods perform faster than the baseline methods. The conclusion hold across three cases with different pairs of $(p, n_i)$, indicating that the advantage of our methods do not change by working on graphs and datasets of different sizes. In addition, when $p = 4000$, JGL-group and JGL-groupinf cannot handle $K \geq 5$ (i.e., the R program died) due to the memory issue on our experiment machine, while both FASJEM-G and FASJEM-I can. This proves that FASJEM requires a lower memory cost than the base-lines. In all subfigures (b), (d) and (f), FASJEM curves are roughly linear. The experimental results match with the computational complexity analysis we have performed in Table 2 (the computation cost of FASJEM is linear to K). Moreover, subfigures (b), (d) and (f) show that both FASJEM-G-GPU and FASJEM-I-GPU implementations spend only $\frac{1}{10}$ time of their CPU implementations respectively. This confirms that GPU-parallelization can speed up FASJEM significantly.

Furthermore in Section S:6, we compare FASJEM-G and JGL-group on four different real-world datasets. FASJEM-G consistently outperforms JGL-group on all four datasets in recovering more known edges.

# Appendix: A Fast and Scalable Joint Estimator for Learning Multiple Related Sparse Gaussian Graphical Models

**Beilun Wang**
University of Virginia

**Ji Gao**
University of Virginia

**Yanjun Qi**
University of Virginia

## S:1 Appendix: Backward mapping for M-Estimator

The graphical model MLE can be expressed as a backward mapping[1] in an exponential family distribution that computes the model parameters corresponding to some given (sample) moments. There are however two caveats with this backward mapping: it is not available in closed form for many classes of models, and even if it were available in closed form, it need not be well-defined in high-dimensional settings (i.e., could lead to unbounded model parameter estimates).

We provide detailed explanations about backward mapping from the M-estimator framework [2] and backward mapping for Gaussian special case in this section.

**Backward mapping:** Suppose a random variable $X \in \mathbb{R}^p$ follows the exponential family distribution:

$$\mathbb{P}(X;\theta) = h(X)\exp\{<\theta, \phi(\theta)> -A(\theta)\} \quad \text{(S:1–1)}$$

Where $\theta \in \Theta \subset \mathbb{R}^d$ is the canonical parameter to be estimated and $\Theta$ denotes the parameter space, $\phi(X)$ denotes the sufficient statistics with a feature mapping function $\phi : \mathbb{R}^p \to \mathbb{R}^d$, and $A(\theta)$ is the log-partition function. We define mean parameters as: $\nu(\theta) := \mathbb{E}[\phi(X)]$, which are the first moments of the sufficient statistics $\phi(\theta)$ under the exponential family distribution. The set of all possible moments by the moment polytope:

$$\mathcal{M} = \{\nu | \exists p \text{ is a distribution s.t. } \mathbb{E}_p[\phi(X)] = \nu\} \quad \text{(S:1–2)}$$

Most machine learning problem about graphical model inference involves the task of computing moments $\nu(\theta) \in \mathcal{M}$ given the canonical parameters $\theta \in \widehat{H}$. We denote this computing as **forward mapping**:

$$\mathcal{A} : \widehat{H} \to \mathcal{M} \quad \text{(S:1–3)}$$

When we need to consider the reverse computing of the forward mapping, we denote the interior of $\mathcal{M}$ as $\mathcal{M}^0$. The so-called **backward mapping** is defined as:

$$\mathcal{A}^* : \mathcal{M}^0 \to \widehat{H} \quad \text{(S:1–4)}$$

which does not need to be unique. For the exponential family distribution,

$$\mathcal{A}^* : \nu(\theta) \to \theta = \nabla A^*(\nu(\theta)). \quad \text{(S:1–5)}$$

Where $A^*(\nu(\theta)) = \sup_{\theta \in \widehat{H}} <\theta, \nu(\theta)> -A(\theta)$.

**Backward Mapping: Gaussian Case** If the random variable $X \in \mathbb{R}^p$ follows the Gaussian Distribution $N(\mu, \Sigma)$. Then $\theta = (\Sigma^{-1}\mu, -\frac{1}{2}\Sigma^{-1})$. The sufficient statistics $\phi(X) = (X, XX^T)$ and the log-partition function $A(\theta) = \frac{1}{2}\mu^T\Sigma^{-1}\mu + \frac{1}{2}\log(|\Sigma|)$. $h(x) = (2\pi)^{-\frac{k}{2}}$.

When inferring the Gaussian Graphical Models, it is easy to estimate the mean vector $\nu(\theta)$, since it equals to $\mathbb{E}[X, XX^T]$.

Because the $\theta$ contains entry $\Sigma^{-1}$, when estimating sGGM, we need to use the backward mapping:

For the case of Gaussian distribution,

$$\begin{aligned}
\theta &= (\Sigma^{-1}\mu, -\frac{1}{2}\Sigma^{-1}) = \mathcal{A}^*(\nu) = \nabla A^*(\nu) \\
&= ((\mathbb{E}_\theta[XX^T] - \mathbb{E}_\theta[X]\mathbb{E}_\theta[X]^T)^{-1}\mathbb{E}_\theta[X], \quad \text{(S:1–6)} \\
&\quad -\frac{1}{2}(\mathbb{E}_\theta[XX^T] - \mathbb{E}_\theta[X]\mathbb{E}_\theta[X]^T)^{-1}).
\end{aligned}$$

By plugging in $A(\theta) = \frac{1}{2}\mu^T\Sigma^{-1}\mu + \frac{1}{2}\log(|\Sigma|)$ into Eq. (S:1–5), $\Omega$ is canonical parameter using backward mapping. We get $\Omega$ as $(\mathbb{E}_\theta[XX^T] - \mathbb{E}_\theta[X]\mathbb{E}_\theta[X]^T)^{-1}) = \Sigma^{-1}$, which can be inferred by the estimated covariance matrix.

## S:2 Appendix: Method and Optimization

**More about Proximal Optimization:** The proximal algorithm only needs to calculate the proximity operator of the parameters to be optimized. The proximity operator in proximal algorithms is defined as:

$$\text{prox}_{\gamma f}(x) = \operatorname*{argmin}_{y}(f(y) + (\frac{1}{2\gamma}||x - y||_2^2)). \quad \text{(S:2–1)}$$

The benefit of the proximal algorithm is that many proximity operators are entry-wise operators for the targeted parameters. The parallel proximal (initially called proximity splitting) algorithm [3] belongs to the general family of distributed convex optimization that optimizes in such a way that each term (in this case, each proximity operator) can be handled by its own processing element, such as a thread or processor.

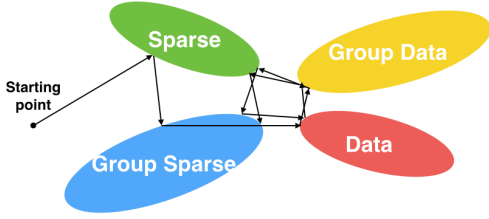

Figure S:1: A simple figure to show how our optimization method works. Our optimization approach is a method with linear convergence rate in finding the optimal point. It considers four properties : (1) information from the raw data; (2) information from the group data; (3)sparsity property; (4) group sparsity property.

**More about four proximity operators for CPU implementation of FASJEM-G:** In the following, we denote $x = \Omega_{tot}$, $a = \Sigma_{tot}$ and $g \in \mathcal{G}$ to simply notations. Eq. (S:2–2) and Eq. (S:2–4) are entry-wise operators and Eq. (S:2–3) and Eq. (S:2–5) are group entry-wise. Group entry-wise means in calculation, the operator can compute each group of entries independently from other groups. Entry-wise means the calculation of each entry is only related to itself). The optimization process of Algorithm 1 iterating among four proximal operators is visualized by Figure S:1.

For $f_1(\cdot) = ||\cdot||_1$.

$$\text{prox}_{\gamma f_1}(x) = \text{prox}_{\gamma ||\cdot||_1}(x)$$
$$= \begin{cases} x_{j,k}^{(i)} - \gamma, & x_{j,k}^{(i)} > \gamma \\ 0, & |x_{j,k}^{(i)}| \leq \gamma \\ x_{j,k}^{(i)} + \gamma, & x_{j,k}^{(i)} < -\gamma \end{cases} \quad \text{(S:2–2)}$$

Eq. (S:2–2) is the closed form solution of Eq. (S:2–1) when $f = |\cdot|_1$. Here $j, k = 1, \ldots, p$ and $i = 1, \ldots, K$. This is an entry-wise operator (i.e., the calculation of each entry is only related to itself).

Similarly, $f_2(\cdot) = ||\cdot||_{\mathcal{G},2}$

$$\text{prox}_{\gamma f_2}(x_g) = \text{prox}_{\gamma ||\cdot||_{\mathcal{G},2}}(x_g)$$
$$= \begin{cases} x_g - \gamma \frac{x_g}{||x_g||_2}, & ||x_g||_2 > \gamma \\ 0, & ||x_g||_2 \leq \gamma \end{cases} \quad \text{(S:2–3)}$$

Here $g \in \mathcal{G}$. This is a group entry-wise operator (computing a group of entries is not related to other groups).

$f_3(\cdot)$ and $f_4(\cdot)$ include function forms of $\mathcal{I}_{f(\cdot)<D}$ and $\text{prox}_{\mathcal{I}_{\{f(\cdot)<D\}}} = \text{proj}_{\{f(\cdot)<D\}}$, where $\text{proj}_C$ means the projection function to the convex set $C$. We can obtain

$$\text{prox}_{\gamma f_3}(x) = \text{proj}_{||x-a||_\infty \leq \lambda}$$
$$= \begin{cases} x_{j,k}^{(i)}, & |x_{j,k}^{(i)} - a_{j,k}^{(i)}| \leq \lambda \\ a_{j,k}^{(i)} + \lambda, & x_{j,k}^{(i)} > a_{j,k}^{(i)} + \lambda \\ a_{j,k}^{(i)} - \lambda, & x_{j,k}^{(i)} < a_{j,k}^{(i)} - \lambda \end{cases} \quad \text{(S:2–4)}$$

where $j, k = 1, \ldots, p$ and $i = 1, \ldots, K$. This operator is entry-wise (i.e., only related to each entry of $x$ and $a$).

$$\text{prox}_{\gamma f_4}(x_g) = \text{proj}_{||x-a||_{\mathcal{G},2}^* \leq \lambda}$$
$$= \begin{cases} x_g, & ||x_g - a_g||_2 \leq \lambda \\ \lambda \frac{x_g - a_g}{||x_g - a_g||_2} + a_g, & ||x_g - a_g||_2 > \lambda \end{cases} \quad \text{(S:2–5)}$$

This operator is group entry-wise.

**More about four proximity operators for GPU parallel implementation of FASJEM-G:** The four proximity operators on GPU are summarized in Table 1. More details as following:

For Eq. (S:2–2),

$$\text{prox}_{\gamma f_1}(x) = \text{prox}_{\gamma ||\cdot||_1}(x)$$
$$= \max((x_{j,k}^{(i)} - \gamma), 0) + \min(0, (x_{j,k}^{(i)} + \gamma)) \quad \text{(S:2–6)}$$

For Eq. (S:2–3)

$$\text{prox}_{\gamma f_2}(x_g) = \text{prox}_{\gamma ||\cdot||_{\mathcal{G},2}}(x_g)$$
$$= x_g \max((1 - \frac{\gamma}{||x_g||_2}), 0) \quad \text{(S:2–7)}$$

For Eq. (S:2–4)

$$\text{prox}_{\gamma f_3}(x) = \text{proj}_{||x-a||_\infty \leq \lambda}$$
$$= \min(\max(x_{j,k}^{(i)} - a_{j,k}^{(i)}, -\lambda), \lambda) + a_{j,k}^{(i)} \quad \text{(S:2–8)}$$

For Eq. (S:2–5)

$$\text{prox}_{\gamma f_4}(x) = \text{proj}_{||x-a||_{\mathcal{G},2}^* \leq \lambda}$$
$$= \max(\frac{\lambda}{||x_g - a_g||_2}, 1)(x_g - a_g) + a_g \quad \text{(S:2–9)}$$

Here $j, k = 1, \ldots, p$, $i = 1, \ldots, K$ and $g \in \mathcal{G}$.

**More about Q-linearly Convergence of Optimization:** The proposed optimization is a first-order method. Based on the recent study[4], the optimization sequence $\{\Omega^i\}$ (for $i = 1$ to $t$ iteration) converges Q-linearly. Q-linearly means:

$$\limsup_{k \to \infty} \frac{||\Omega^{k+1} - \Omega^*||}{||\Omega^k - \Omega^*||} \leq \rho \quad \text{(S:2–10)}$$

## S:3 Appendix: Related previous studies using elementary based estimators

Related previous studies based on elementary estimators are summarized in Table S:1.

## S:4 Appendix: More about Experimental Setting and Baselines

**Hyperparameter tuning:** We have tried BIC method (used in [5]) for choosing the tuning parameter $\lambda_n$. As pointed out by ([6], [7] and [8]), the BIC or AIC method may not work well for the high-dimensional case. Therefore we have skipped adding the results from BIC or AIC.

In our experiments, we compare our model with the baselines by varying the same set of the tuning parameters.

**Baseline:** Recent literature[9] shows that the single sGGM has a close form solution through the EE estimator (i.e., no iteration). It is not fair to compare our estimator to such a closed-form sGGM estimator in terms of the speed or memory usage. Therefore we don't include the single sGGM as a baseline.

**Real World Experiments:** We also tried FASJEM-I and JGL-groupinf on the three datasets. No matched interactions were found in one dataset. Therefore, we omit the results.

## S:5 Appendix: More Experimental Results from Simulated Data

Figure S:3 represents a comparison between the single-task EE estimator for sGGM and GLasso estimator. We choose the $\Omega^{(i)}$ in the random graph model as the true graph. We obtain the two subfigures by varying $p$ in a set of $\{100, 200, 300, 400, 500\}$. The left subfigure is "AUC vs. p (number of features)" while the right subfigure is "Time vs. p (number of features)". Figure S:3 shows that the elementary estimator has achieved similar performance of GLasso among different $p$ while the computation time of EE is much less than the GLasso.

## S:6 Experiments on Real-world Datasets

We apply FASJEM-G and JGL-group on four different real-world datasets: (1) the breast/colon cancer data [10] (with 2 cell types and 104 samples, each having 22283 features); (2) Crohn's disease data [11] ( with 3 cell types, 127 samples and 22283 features) , (3) the myeloma and bone lesions data set[12] (with 2 cell types, 173 samples and 12625 features) and (4) Encode project dataset[13] (with 3 cell types, 25185 samples and 27 features). For the first three datasets, we select its top 500 features based on the variance of the variables. After obtaining estimated dependency networks, we compare all methods using two major existing databases [14, 15] archiving known gene interactions. The number of known gene-gene interactions predicted by each method has been shown as bar graphs in Figure S:4. These graphs clearly show that FASJEM-G outperforms JGL-group on all three datasets and across all cell conditions within each of the three datasets. This leads us to believe that the proposed FASJEM-G is very promising for identifying variable interactions in a wider range of applications as well.

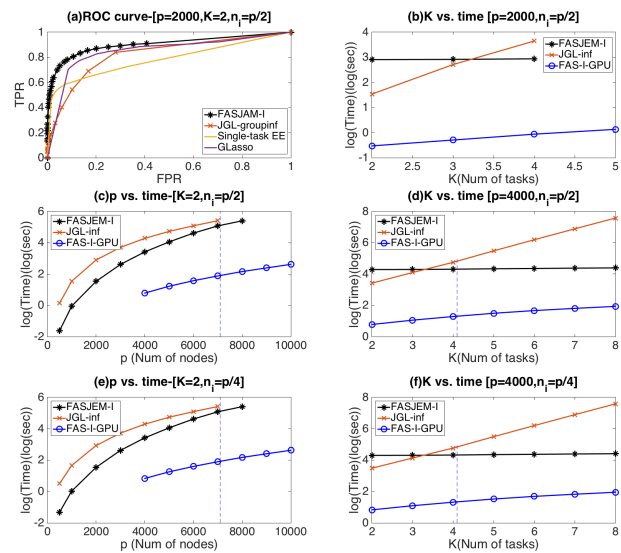

Figure S:2: Comparison between FASJEM-I and JGL-groupinf using accuracy, speed and memory capacity. (a) FPR-TPR curves of two methods on the simulated dataset using Random Graph Model when $p = 2000$ and $K = 2$. (c) and (e) Time versus $p$(the number of variables) curves from FASJEM-G, JGL-group and FASJEM-I's GPU implementation. (c) uses $n_i = p/2$ and (e)$n_i = p/4$ (b), (d) and (f) include the time versus $K$(the number of tasks) curves for two methods plus FASJEM-I-GPU. (b) uses $p = 2000$ and $n_i = p/2$, (d) uses $p = 4000$ and $n_i = p/2$ and (f) uses $p = 4000$ and $n_i = p/4$.

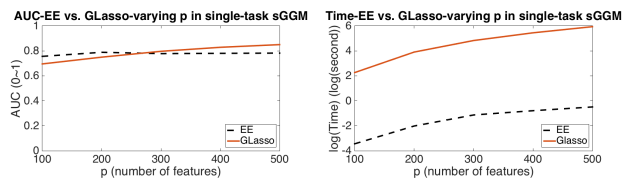

Figure S:3: Comparison between elementary estimator for sGGM and GLasso for single-task sGGM. The left figure is the curve of AUC number by varying $p$. The number of sample $n = p/2$. The right figure is the curve of computation time by varying $p$. Other settings are the same as the left one. Clearly, elementary estimator has the similar accuracy performance as GLasso but is much faster and scalable than it.

Table S:1: Two categories of relevant studies differ over learning based on "penalized log-likelihood" or learning based on "elementary estimator"

| Problems | Penalized Likelihood | Elementary estimator |
|---|---|---|
| High dimensional linear regression | Lasso: $\operatorname*{argmin}_{\beta} \|Y-\beta X\|_F + \lambda\|\beta\|_1$ | $\operatorname*{argmin}_{\beta} \|\beta\|_1$ subject to: $\|\beta - (X^TX + \epsilon I)^{-1}X^Ty\|_\infty \leq \lambda_n$ |
| sparse Gaussian Graphical Model | gLasso: $\operatorname*{argmin}_{\Omega \geq 0} -logdet(\Omega) + <\Omega, \Sigma> + \lambda\|\Omega\|_1$ | $\operatorname*{argmin}_{\Omega \geq 0} \|\Omega\|_1$ subject to: $\|\Omega - [T_v(\Sigma)]^{-1}\|_\infty \leq \lambda_n$ |
| Multi-task sGGM | Different Choices for Penalty $\mathcal{R}'$ $\operatorname*{argmin}_{\Omega > 0} \sum_i (-L(\Omega_{tot}) + \lambda_1 \sum_i \|\Omega^{(i)}\|_1 + \lambda_2 \mathcal{R}'(\Omega_{tot})$ | **Our method: FASJEM** |

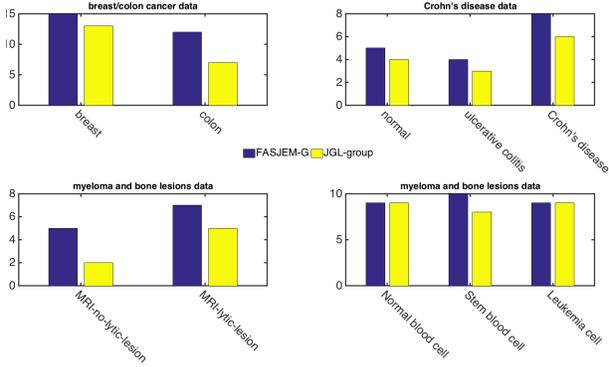

Figure S:4: Compare predicted dependencies among genes or proteins using existing databases [14, 15] with known interactions (biologically validated) in human. The number of matches among predicted interactions and known interactions is shown as bar lines.

## S:7 Appendix: More about the theoretical error bounds

**Background–error bound for elementary estimator:** For proving the error bounds, we first briefly review the error bound of a single-task EE-based model using the proof strategy in the unified framework[2]. The single task-EE follows the general formulation:

$$\operatorname*{argmin}_{\theta} \mathcal{R}(\theta)$$
$$\text{subject to:} \mathcal{R}^*(\widehat{\theta}_n - \theta) \leq \lambda_n \quad \text{(S:7–1)}$$

where $\mathcal{R}(\cdot)$ is the $\ell_1$ regularization function and $\widehat{\theta}_n$ is the backforward mapping for $\theta$.

Following the same proof strategy in the unified framework [2], we first decompose the parameter space into a subspace pair $(\mathcal{M}, \bar{\mathcal{M}}^\perp)$, where $\bar{\mathcal{M}}$ is the closure of $\mathcal{M}$. Here $\mathcal{M}$ is the **model subspace** that typically has a much lower dimension than the original high-dimensional space. $\bar{\mathcal{M}}^\perp$ is the **perturbation subspace** of parameters. For further proofs, we assume the regularization function in Eq. (S:7–1) is **decomposable** w.r.t the subspace pair $(\mathcal{M}, \bar{\mathcal{M}}^\perp)$.

**(C1)** $\mathcal{R}(u+v) = \mathcal{R}(u) + \mathcal{R}(v), \forall u \in \mathcal{M}, \forall v \in \bar{\mathcal{M}}^\perp$.

[2] shows that most regularization norms are decomposable corresponding to a certain subspace pair.

**Definition S:7.1.** *A term **subspace compatibility constant** is defined as* $\Psi(\mathcal{M}, |\cdot|) := \sup_{u \in \mathcal{M}\backslash\{0\}} \frac{\mathcal{R}(u)}{|u|}$ *which captures the relative value between the error norm $|\cdot|$ and the regularization function $\mathcal{R}(\cdot)$.*

For simplicity, we assume there exists a true parameter $\theta^*$ which has the exact structure w.r.t a certain subspace pair. That is:

**(C2)** $\exists$ a subspace pair $(\mathcal{M}, \bar{\mathcal{M}}^\perp)$ such that the true parameter satisfies $\operatorname{proj}_{\mathcal{M}^\perp}(\theta^*) = 0$

Then we have the following theorem.

**Theorem S:7.2.** *Suppose the regularization function in Eq. (S:7–1) satisfies condition **(C1)**, the true parameter of Eq. (S:7–1) satisfies condition **(C2)**, and $\lambda_n$ satisfies that $\lambda_n \geq \mathcal{R}^*(\widehat{\theta} - \theta^*)$. Then, the optimal solution $\widehat{\theta}$ of Eq. (S:7–1) satisfies:*

$$\mathcal{R}^*(\widehat{\theta} - \theta^*) \leq 2\lambda_n \quad \text{(S:7–2)}$$

$$\|\widehat{\theta} - \theta^*\|_2 \leq 4\lambda_n \Psi(\bar{\mathcal{M}}) \quad \text{(S:7–3)}$$

$$\mathcal{R}(\widehat{\theta} - \theta^*) \leq 8\lambda_n \Psi(\bar{\mathcal{M}})^2 \quad \text{(S:7–4)}$$

## S:8 Proof

**Proof of Theorem (S:7.2)**

*Proof.* Let $\Delta := \widehat{\theta} - \theta^*$ be the error vector that we are interested in.

$$\begin{aligned}\mathcal{R}^*(\widehat{\theta} - \theta^*) &= \mathcal{R}^*(\widehat{\theta} - \widehat{\theta}_n + \widehat{\theta}_n - \theta^*) \\ &\leq \mathcal{R}^*(\widehat{\theta}_n - \widehat{\theta}) + \mathcal{R}^*(\widehat{\theta}_n - \theta^*) \leq 2\lambda_n\end{aligned} \quad \text{(S:8–1)}$$

By the fact that $\theta^*_{\mathcal{M}^\perp} = 0$, and the decomposability of $\mathcal{R}$ with respect to $(\mathcal{M}, \bar{\mathcal{M}}^\perp)$

$$\begin{aligned}
&\mathcal{R}(\theta^*) \\
&= \mathcal{R}(\theta^*) + \mathcal{R}[\Pi_{\bar{\mathcal{M}}^\perp}(\Delta)] - \mathcal{R}[\Pi_{\bar{\mathcal{M}}^\perp}(\Delta)] \\
&= \mathcal{R}[\theta^* + \Pi_{\bar{\mathcal{M}}^\perp}(\Delta)] - \mathcal{R}[\Pi_{\bar{\mathcal{M}}^\perp}(\Delta)] \\
&\leq \mathcal{R}[\theta^* + \Pi_{\bar{\mathcal{M}}^\perp}(\Delta) + \Pi_{\bar{\mathcal{M}}}(\Delta)] + \mathcal{R}[\Pi_{\bar{\mathcal{M}}}(\Delta)] \\
&\quad - \mathcal{R}[\Pi_{\bar{\mathcal{M}}^\perp}(\Delta)] \\
&= \mathcal{R}[\theta^* + \Delta] + \mathcal{R}[\Pi_{\bar{\mathcal{M}}}(\Delta)] - \mathcal{R}[\Pi_{\bar{\mathcal{M}}^\perp}(\Delta)]
\end{aligned}$$
(S:8–2)

Here, the inequality holds by the triangle inequality of norm. Since Eq. (S:7–1) minimizes $\mathcal{R}(\widehat{\theta})$, we have $\mathcal{R}(\theta^* + \Delta) = \mathcal{R}(\widehat{\theta}) \leq \mathcal{R}(\theta^*)$. Combining this inequality with Eq. (S:8–2), we have:

$$\mathcal{R}[\Pi_{\bar{\mathcal{M}}^\perp}(\Delta)] \leq \mathcal{R}[\Pi_{\bar{\mathcal{M}}}(\Delta)] \quad \text{(S:8–3)}$$

Moreover, by Hölder's inequality and the decomposability of $\mathcal{R}(\cdot)$, we have:

$$\begin{aligned}
||\Delta||_2^2 &= \langle \Delta, \Delta \rangle \leq \mathcal{R}^*(\Delta)\mathcal{R}(\Delta) \leq 2\lambda_n \mathcal{R}(\Delta) \\
&= 2\lambda_n[\mathcal{R}(\Pi_{\bar{\mathcal{M}}}(\Delta)) + \mathcal{R}(\Pi_{\bar{\mathcal{M}}^\perp}(\Delta))] \leq 4\lambda_n \mathcal{R}(\Pi_{\bar{\mathcal{M}}}(\Delta)) \\
&\leq 4\lambda_n \Psi(\bar{\mathcal{M}})||\Pi_{\bar{\mathcal{M}}}(\Delta)||_2
\end{aligned}$$
(S:8–4)

where $\Psi(\bar{\mathcal{M}})$ is a simple notation for $\Psi(\bar{\mathcal{M}}, ||\cdot||_2)$.

Since the projection operator is defined in terms of $||\cdot||_2$ norm, it is non-expansive: $||\Pi_{\bar{\mathcal{M}}}(\Delta)||_2 \leq ||\Delta||_2$. Therefore, by Eq. (S:8–4), we have:

$$||\Pi_{\bar{\mathcal{M}}}(\Delta)||_2 \leq 4\lambda_n \Psi(\bar{\mathcal{M}}), \quad \text{(S:8–5)}$$

and plugging it back to Eq. (S:8–4) yields the error bound Eq. (S:7–3).

Finally, Eq. (S:7–4) is straightforward from Eq. (S:8–3) and Eq. (S:8–5).

$$\begin{aligned}
\mathcal{R}(\Delta) &\leq 2\mathcal{R}(\Pi_{\bar{\mathcal{M}}}(\Delta)) \\
&\leq 2\Psi(\bar{\mathcal{M}})||\Pi_{\bar{\mathcal{M}}}(\Delta)||_2 \leq 8\lambda_n \Psi(\bar{\mathcal{M}})^2.
\end{aligned}$$
(S:8–6)

□

**Proof of Theorem (5.3)**

*Proof.* In this proof, we consider the matrix parameter such as the covariance. $I = \{1, 2\}$ in the following contents. Basically, the Frobenius norm can be simply replaced by $\ell_2$ norm for the vector parameters. Let $\Delta_i := \widehat{\theta}_i - \theta^*_i$, and $\Delta = \widehat{\theta} - \theta^* = \Sigma_{i \in I}\Delta_i$. The error bound Eq. (5.3) can be easily shown from the assumption in the statement with the constraint of Eq. (5.2). For every $i \in I$,

$$\begin{aligned}
\mathcal{R}^*_i(\Delta) &= \mathcal{R}^*_i(\widehat{\theta} - \theta^*) = \mathcal{R}^*_i(\widehat{\theta} - \widehat{\theta}_n + \widehat{\theta}_n - \theta^*) \\
&\leq \mathcal{R}^*_i(\widehat{\theta}_n - \widehat{\theta}) + \mathcal{R}^*_i(\widehat{\theta}_n - \theta^*) \leq 2\lambda_i.
\end{aligned}$$
(S:8–7)

By the similar reasoning as in Eq. (S:8–2) with the fact that $\Pi_{\mathcal{M}^\perp_i}(\theta^*_i) = 0$ in **C3**, and the decomposability of $\mathcal{R}_i$ with respect to $(\mathcal{M}_i, \widehat{\mathcal{M}}^\perp_i)$, we have:

$$\begin{aligned}
\mathcal{R}_i(\theta^*_i) \leq &\mathcal{R}_i[\theta^*_i + \Delta_i] + \mathcal{R}_i[\Pi_{\bar{\mathcal{M}}_i}(\Delta_i)] \\
&- \mathcal{R}_i[\Pi_{\bar{\mathcal{M}}^\perp_i}(\Delta_i)].
\end{aligned}$$
(S:8–8)

Since $\left\{\widehat{\theta}_i\right\}_{i \in I}$ minimizes the objective function of Eq. (5.2),

$$\begin{aligned}
\sum_{i \in I} \lambda_i \mathcal{R}_i(\widehat{\theta}_i) \leq \sum_{i \in I} \lambda_i \{&\mathcal{R}_i(\theta^*_i + \Delta_i) \\
&\mathcal{R}_i[\Pi_{\bar{\mathcal{M}}_i}(\Delta_i)] - \mathcal{R}_i[\Pi_{\bar{\mathcal{M}}^\perp_i}(\Delta_i)]\},
\end{aligned}$$
(S:8–9)

Which implies

$$\sum_{i \in I} \lambda_i \mathcal{R}_i[\Pi_{\bar{\mathcal{M}}^\perp_i}(\Delta_i)] \leq \sum_{i \in I} \lambda_i \mathcal{R}_i[\Pi_{\bar{\mathcal{M}}_i}(\Delta_i)] \quad \text{(S:8–10)}$$

Now, for each structure $i \in I$, we have an application for Hölder's inequality: $|\langle\langle \Delta, \Delta_i \rangle\rangle| \leq \mathcal{R}^*_i(\Delta)\mathcal{R}_i(\Delta_i) \leq 2\lambda_i \mathcal{R}_i(\Delta_i)$ where the notation $\langle\langle A, B \rangle\rangle$ denotes the trace inner product, $\text{trace}(A^T B) = \Sigma_i \Sigma_j A_{ij}B_{ij}$, and we use the pre-computed bound in Eq. (S:8–7). Then, the Frobenius error $||\Delta||_F$ can be upper-bounded as follows:

$$\begin{aligned}
||\Delta||_F^2 &= \langle\langle \Delta, \Delta \rangle\rangle = \sum_{i \in I}\langle\langle \Delta, \Delta_i \rangle\rangle \leq \sum_{i \in I}|\langle\langle \Delta, \Delta_i \rangle\rangle| \\
&\leq 2\sum_{i \in I}\lambda_i \mathcal{R}_i(\Delta_i) \leq 2\sum_{i \in I}\{\lambda_i \mathcal{R}_i[\Pi_{\bar{\mathcal{M}}_i}(\Delta_i)] + \\
&\lambda_i \mathcal{R}_i[\Pi_{\bar{\mathcal{M}}^\perp_i}(\Delta_i)]\} \leq 4\sum_{i \in I}\lambda_i \mathcal{R}_i[\Pi_{\bar{\mathcal{M}}_i}(\Delta_i)] \\
&\leq 4\sum_{i \in I}\lambda_i \Psi(\bar{\mathcal{M}}_i)||\Pi_{\bar{\mathcal{M}}_i}(\Delta_i)||_F
\end{aligned}$$
(S:8–11)

where $\Psi(\bar{\mathcal{M}}_i)$ denotes the compatibility constant of space $\bar{\mathcal{M}}_i$ with respect to the Frobenius norm: $\Psi(\bar{\mathcal{M}}_i, ||\cdot||_F)$.

Here, we define a key notation in the error bound:

$$\Phi := \max_{i \in I} \lambda_i \Psi(\bar{\mathcal{M}}_i). \qquad (S:8\text{--}12)$$

Armed with this notation, Eq. (S:8–11) can be written as

$$||\Delta||_F^2 \leq 4\Phi \sum_{i \in I} ||\Pi_{\bar{\mathcal{M}}_i}(\Delta_i)||_F \qquad (S:8\text{--}13)$$

At this point, we directly appeal to the result in Proposition 2 of [16] with a small modification:

**Proposition 4.** Suppose that the structural incoherence condition **(C4)** as well as the condition **(C3)** hold. Then, we have

$$2|\sum_{i<j} \langle\langle \Delta_i, \Delta_j \rangle\rangle| \leq \frac{1}{2} \sum_{i \in I} ||\Delta_i||_F^2. \qquad (S:8\text{--}14)$$

By this proposition, we have

$$\begin{aligned}
\sum_{i \in I} ||\Delta_i||_F^2 &\leq ||\Delta||_F^2 + 2|\sum_{i<j} \langle\langle \Delta_i, \Delta_j \rangle\rangle| \\
&\leq ||\Delta||_F^2 + \frac{1}{2} \sum_{i \in I} ||\Delta_i||_F^2,
\end{aligned} \qquad (S:8\text{--}15)$$

which implies $\Sigma_{i \in I} ||\Delta_i||_F^2 \leq 2||\Delta||_F^2$.

Moreover, since the projection operator is defined in terms of the Frobenius norm, it is non-expansive for all $i$: $||\Pi_{\bar{\mathcal{M}}_i}(\Delta_i)||_F \leq ||\Delta_i||_F$. Hence, we finally obtain:

$$\begin{aligned}
(\sum_{i \in I} ||\Pi_{\bar{\mathcal{M}}_i}(\Delta_i)||_F)^2 &\leq (\sum_{i \in I} ||\Delta_i||_F)^2 \\
&\leq |I| \sum_{i \in I} ||\Delta_i||_F^2 \leq 8|I|\Phi \sum_{i \in I} ||\Pi_{\bar{\mathcal{M}}_i}(\Delta_i)||_F
\end{aligned} \qquad (S:8\text{--}16)$$

and therefore,

$$\sum_{i \in I} ||\Pi_{\bar{\mathcal{M}}_i}(\Delta_i)||_F \leq 8|I|\Phi \qquad (S:8\text{--}17)$$

The Frobenius norm error bound Eq. (5.5) can be derived by plugging Eq. (S:8–17) back into Eq. (S:8–13):

$$||\Delta||_F^2 \leq 32|I|\Phi^2. \qquad (S:8\text{--}18)$$

Therefore, we have

$$||\Delta||_F \leq 8\Phi \qquad (S:8\text{--}19)$$

Which is exactly Eq. (5.5)

The proof of the final error bound Eq. (5.4) is straightforward from Eq. (S:8–10) and Eq. (S:8–17) as follows: for each fixed $i \in I$,

$$\begin{aligned}
&\mathcal{R}_i(\Delta_i) \\
&\leq \frac{1}{\lambda_i}\{\lambda_i \mathcal{R}_i[\Pi_{\bar{\mathcal{M}}_i}(\Delta_i)] + \lambda_i \mathcal{R}_i[\Pi_{\bar{\mathcal{M}}_i^\perp}(\Delta_i)]\} \\
&\leq \frac{1}{\lambda_i}\{\lambda_i \mathcal{R}_i[\Pi_{\bar{\mathcal{M}}_i}(\Delta_i)] + \sum_{j \in I} \lambda_j \mathcal{R}_j[\Pi_{\bar{\mathcal{M}}_j}(\Delta_j)]\} \\
&\leq \frac{2}{\lambda_i} \sum_{j \in I} \lambda_j \mathcal{R}_j[\Pi_{\bar{\mathcal{M}}_j}(\Delta_j)] \\
&\leq \frac{2}{\lambda_i} \sum_{j \in I} \lambda_j \Psi(\bar{\mathcal{M}}_j) ||\Pi_{\bar{\mathcal{M}}_j}(\Delta_j)||_F \\
&\leq \frac{2\Phi}{\lambda_i} \sum_{j \in I} ||\Pi_{\bar{\mathcal{M}}_j}(\Delta_j)||_F \leq \frac{16|I|\Phi^2}{\lambda_i} = \frac{32\Phi^2}{\lambda_i}
\end{aligned} \qquad (S:8\text{--}20)$$

which completes the proof. $\square$

**Proof of Theorem (5.4)**

*Proof.* Since $\lambda_n > \lambda_n'$ and $\sqrt{s} > \sqrt{s_\mathcal{G}}$, We have that $\lambda_n \sqrt{s} > \lambda_n' \sqrt{s_\mathcal{G}}$.

By Theorem (5.3),

$||\widehat{\Omega}_{tot} - \Omega_{tot}^*||_F \leq 8\max(\lambda_n\sqrt{s}, \lambda_n'\sqrt{s_\mathcal{G}}) \leq 8\sqrt{s}\lambda_n$. $\square$

### S:8.1  Useful lemma(s)

**Lemma S:8.1.** *(Theorem 1 of [17]). Let $\delta$ be $\max_{ij}|[\frac{X^T X}{n}]_{ij} - \Sigma_{ij}|$. Suppose that $\nu > 2\delta$. Then, under the conditions (C-Sparse$\Sigma$), and as $\rho_v(\cdot)$ is a soft-threshold function, we can deterministically guarantee that the spectral norm of error is bounded as follows:*

$$|||T_v(\widehat{\Sigma}) - \Sigma|||_\infty \leq 5\nu^{1-q}c_0(p) + 3\nu^{-q}c_0(p)\delta \quad (S:8\text{--}21)$$

**Lemma S:8.2.** *(Lemma 1 of [18]). Let $\mathcal{A}$ be the event that*

$$||\frac{X^T X}{n} - \Sigma||_\infty \leq 8(\max_i \Sigma_{ii})\sqrt{\frac{10\tau \log p'}{n}} \quad (S:8\text{--}22)$$

*where $p' := \max n, p$ and $\tau$ is any constant greater than 2. Suppose that the design matrix $X$ is i.i.d. sampled*

*from $\Sigma$-Gaussian ensemble with $n \geq 40 \max_i \Sigma_{ii}$. Then, the probability of event $\mathcal{A}$ occurring is at least $1 - 4/p'^{\tau-2}$.*

**Proof of Corollary (5.5)**

*Proof.* In the following proof, we re-denote the following two notations: $\Sigma_{tot} := \begin{pmatrix} \Sigma^{(1)} & 0 & \cdots & 0 \\ 0 & \Sigma^{(2)} & \cdots & 0 \\ \vdots & \vdots & \ddots & \vdots \\ 0 & 0 & \cdots & \Sigma^{(K)} \end{pmatrix}$

and

$\Omega_{tot} := \begin{pmatrix} \Omega^{(1)} & 0 & \cdots & 0 \\ 0 & \Omega^{(2)} & \cdots & 0 \\ \vdots & \vdots & \ddots & \vdots \\ 0 & 0 & \cdots & \Omega^{(K)} \end{pmatrix}$

The condition (C-Sparse$\Sigma$) and condition (C-MinInf$\Sigma$) also hold for $\Omega_{tot}^*$ and $\Sigma_{tot}^*$. In order to utilize Theorem (5.4) for this specific case, we only need to show that $||\Omega_{tot}^* - [T_\nu(\widehat{\Sigma}_{tot})]^{-1}||_\infty \leq \lambda_n$ for the setting of $\lambda_n$ in the statement:

$$\begin{aligned}
||\Omega_{tot}^* - [T_\nu(\widehat{\Sigma}_{tot})]^{-1}||_\infty &= ||[T_\nu(\widehat{\Sigma}_{tot})]^{-1}(T_\nu(\widehat{\Sigma}_{tot})\Omega_{tot}^* - I)||_\infty \\
&\leq ||||T_\nu(\widehat{\Sigma}_{tot})w||||_\infty ||T_\nu(\widehat{\Sigma}_{tot})\Omega_{tot}^* - I||_\infty \\
&= ||||T_\nu(\widehat{\Sigma}_{tot})]^{-1}|||_\infty ||\Omega_{tot}^*(T_\nu(\widehat{\Sigma}_{tot}) - \Sigma_{tot}^*)||_\infty \\
&\leq ||||T_\nu(\widehat{\Sigma}_{tot})]^{-1}|||_\infty |||\Omega_{tot}^*|||_\infty ||T_\nu(\widehat{\Sigma}_{tot}) - \Sigma_{tot}^*||_\infty.
\end{aligned}$$
(S:8–23)

We first compute the upper bound of $||||[T_\nu(\widehat{\Sigma}_{tot})]^{-1}|||_\infty$. By the selection $\nu$ in the statement, Lemma (S:8.1) and Lemma (S:8.2) hold with probability at least $1 - 4/p'^{\tau-2}$. Armed with Eq. (S:8–21), we use the triangle inequality of norm and the condition (C-Sparse$\Sigma$): for any $w$,

$$\begin{aligned}
||T_\nu(\widehat{\Sigma}_{tot})w||_\infty &= ||T_\nu(\widehat{\Sigma}_{tot})w - \Sigma w + \Sigma w||_\infty \\
&\geq ||\Sigma w||_\infty - ||(T_\nu(\widehat{\Sigma}_{tot}) - \Sigma)w||_\infty \\
&\geq \kappa_2 ||w||_\infty - ||(T_\nu(\widehat{\Sigma}_{tot}) - \Sigma)w||_\infty \\
&\geq (\kappa_2 - ||(T_\nu(\widehat{\Sigma}_{tot}) - \Sigma)w||_\infty)||w||_\infty
\end{aligned}$$
(S:8–24)

Where the second inequality uses the condition (C-Sparse$\Sigma$). Now, by Lemma (S:8.1) with the selection of $\nu$, we have

$$|||T_\nu(\widehat{\Sigma}_{tot}) - \Sigma|||_\infty \leq c_1 (\frac{\log p'}{n_{tot}})^{(1-q)/2} c_0(p) \quad \text{(S:8–25)}$$

where $c_1$ is a constant related only on $\tau$ and $\max_i \Sigma_{ii}$. Specifically, it is defined as $6.5(16(\max_i \Sigma_{ii})\sqrt{10\tau})^{1-q}$. Hence, as long as $n_{tot} > (\frac{2c_1 c_0(p)}{\kappa_2})^{\frac{2}{1-q}} \log p'$ as stated, so that $|||T_\nu(\widehat{\Sigma}_{tot}) - \Sigma|||_\infty \leq \frac{\kappa_2}{2}$, we can conclude that $||T_\nu(\widehat{\Sigma}_{tot})w||_\infty \geq \frac{\kappa_2}{2}||w||_\infty$, which implies $||||[T_\nu(\widehat{\Sigma}_{tot})]^{-1}|||_\infty \leq \frac{2}{\kappa_2}$.

The remaining term in Eq. (S:8–23) is $||T_\nu(\widehat{\Sigma}_{tot}) - \Sigma_{tot}^*||_\infty$; $||T_\nu(\widehat{\Sigma}_{tot}) - \Sigma_{tot}^*||_\infty \leq ||T_\nu(\widehat{\Sigma}_{tot}) - \widehat{\Sigma}_{tot}||_\infty + ||\widehat{\Sigma}_{tot} - \Sigma_{tot}^*||_\infty$. By construction of $T_\nu(\cdot)$ in (C-Thresh) and by Lemma (S:8.2), we can confirm that $||T_\nu(\widehat{\Sigma}_{tot}) - \widehat{\Sigma}_{tot}||_\infty$ as well as $||\widehat{\Sigma}_{tot} - \Sigma_{tot}^*||_\infty$ can be upper-bounded by $\nu$.

By combining all together, we can confirm that the selection of $\lambda_n$ satisfies the requirement of Theorem (5.4), which completes the proof. $\square$